\documentclass[10pt,twocolumn,letterpaper]{article}

\usepackage{iccv}              

\definecolor{iccvblue}{rgb}{0.21,0.49,0.74}
\usepackage[pagebackref,breaklinks,colorlinks,allcolors=iccvblue]{hyperref}

\usepackage{amsmath,amsfonts,bm}
\usepackage{mathtools}

\def\eqref#1{equation~\ref{#1}}

\def\1{\bm{1}}

\DeclareMathAlphabet{\mathsfit}{\encodingdefault}{\sfdefault}{m}{sl}
\SetMathAlphabet{\mathsfit}{bold}{\encodingdefault}{\sfdefault}{bx}{n}

\newcommand{\E}{\mathbb{E}}

\newcommand{\Eb}[2]{\E_{#1}\!\left[#2\right]}

\newcommand{\bc}{\mathbf{c}}

\newcommand{\bx}{\mathbf{x}}

\newcommand{\bepsilon}{{\boldsymbol{\epsilon}}}

\usepackage[capitalize]{cleveref}
    \crefname{section}{Sec.}{Secs.}
    \Crefname{section}{Section}{Sections}
    \Crefname{table}{Table}{Tables}
    \crefname{table}{Table}{Tables} 

\usepackage{multirow}
\usepackage{adjustbox}
\usepackage{svg}
\usepackage{amsmath}
\usepackage[normalem]{ulem}
\useunder{\uline}{\ul}{}
\usepackage{graphicx}
\usepackage{amsmath}
\usepackage{tabularx}
\usepackage{xspace}
\usepackage[linesnumbered,ruled,vlined]{algorithm2e}
\usepackage{graphicx}
\usepackage{amssymb}
\usepackage{booktabs}
\usepackage{dsfont}

\def\paperID{10594} 
\def\confName{ICCV}
\def\confYear{2025}

\title{Continual Personalization for Diffusion Models}

\newcommand{\ours}{\textsc{CNS}\xspace}
\newcommand{\general}{\textit{general neurons}\xspace}
\newcommand{\concept}{\textit{concept neurons}\xspace}
\newcommand{\base}{\textit{base neurons}\xspace}

\author{Yu-Chien Liao$^{*1}$\quad Jr-Jen Chen$^{*1}$\quad Chi-Pin Huang$^{1}$\quad Ci-Siang Lin$^{1}$\quad\\
Meng-Lin Wu$^{2}$\quad Yu-Chiang Frank Wang$^{1}$\\
$^{1}$National Taiwan University \qquad
$^{2}$Qualcomm Technologies, Inc. \\
}

\begin{document}

\twocolumn[{%
\maketitle
\vspace{-2.0em}
\renewcommand\twocolumn[1][]{#1}%
    \includegraphics[width=\linewidth]{fig/teaser_v3.pdf}
    \captionof{figure}{\textbf{Continual personalization.} We present \textbf{C}oncept \textbf{N}euron \textbf{S}election, \ours, a simple yet effective approach to incrementally customize visual concepts. By finetuning concept-related neurons, \ours preserves the zero-shot capabilities of pretrained diffusion models and alleviates catastrophic forgetting problems.}
    \label{fig:teaser}
    \vspace{2em}
}]

\begin{abstract}
Updating diffusion models in an incremental setting would be practical in real-world applications yet computationally challenging. We present a novel learning strategy of \textbf{C}oncept \textbf{N}euron \textbf{S}election, a simple yet effective approach to perform personalization in a continual learning scheme. \ours uniquely identifies neurons in diffusion models that are closely related to the target concepts. In order to mitigate catastrophic forgetting problems while preserving zero-shot text-to-image generation ability, \ours finetunes concept neurons in an incremental manner and jointly preserves knowledge learned of previous concepts. Evaluation of real-world datasets demonstrates that \ours achieves state-of-the-art performance with minimal parameter adjustments, outperforming previous methods in both single and multi-concept personalization works. \ours also achieves fusion-free operation, reducing memory storage and processing time for continual personalization.
\label{sec:abstract}
\end{abstract}

\begingroup
\renewcommand\thefootnote{}
\footnotetext{$^{*}$ denotes co-first author.}
\addtocounter{footnote}{-1}
\endgroup
\section{Introduction}
\label{sec:intro}

Latent Diffusion Models (LDMs)~\citep{rombach2022high} represent a milestone for image generation task by leveraging vast collection of text-image pairs and denoising process. LDMs enable users to create high-quality images through simple text prompts. Yet, LDMs often fall short in generating user-specific concepts (\eg their pets, a scene in a national park), which poses a practical challenge in text-to-image generation since these user-specific concepts are hard to describe directly by text. To address this issue, personalization techniques allow users to adapt LDMs to generate their desired specific content by finetuning it with their own examples. To realize single concept personalization, pioneer works first incorporate techniques such as prompt tuning~\citep{gal2022image} or weights finetuning~\citep{ruiz2023dreambooth}. However, when there are multiple concepts to be learned, naively applying these methods to compose multiple concepts in a single image usually results in overfitting and attributes binding~\citep{chefer2023attend, feng2022training, yu2022scaling, lee2023aligning, wu2023human, ma2024directed, jang2024identity}, which means the model fail to correctly separate characteristics of each concept and generate mix-ups images. To address this concern, \citet{kumari2023multi} first aggregates multiple weights of personalized models by a constrained least square formula, which still results in huge performance degradation as the number of models' weights increase for multi-concept personalization. Many previous works~\citep{smith2023continual, yang2024lora, po2024orthogonal, gu2024mix, kong2025omg} try to overcome this issue by reducing the optimized weights with LoRA~\citep{hu2021lora} finetuning. A common approach nowadays is to learn each concept with LoRA weights separately and fuse the weights while generating multiple concepts images.

Existing methods~\citep{po2024orthogonal, gu2024mix, kumari2023multi, kwon2024concept, ding2024freecustom} tend to make the assumption that personalized concepts are fixed, which means that storing all of the personalized model weights and multiple times of fusion are required if users need different numbers of personalized concepts across different images. However, in the realistic application scenario, users' personalized concepts never remain static and will incrementally increase. A practical scenario is that users can continuously assign new concepts to a single diffusion model and additional computation effort while generating new composite images is not required. Furthermore, learning the concepts separately usually results in conflict while fusing concepts together. \citet{yang2024lora} further prove that LoRA fusion methods will encounter concept vanishing and concept confusion even with additional information such as human poses or image layout. 

In this paper, we propose \ours, \textbf{C}oncept \textbf{N}euron \textbf{S}election, which is able to identify the neurons relevant to personalized target concepts in an incremental fashion. Specifically, CNS allows one to automatically identify the neurons related to (few-shot) images of a concept (i.e., \textit{base neurons}) and those related to the general image synthesis (i.e., \textit{general neurons)} using diffusion models. By excluding general neurons from the base ones, the \textit{concept neurons} describing the input concept of interest can be selected. Moreover, in order to achieve continual learning, an incremental finetuning scheme with such concept neurons is also proposed. Different from existing continual learning methods~\citep{dong2024continually, smith2023continual}, we do not require to train and store any extra LoRAs when handling multiple concepts. As a result, our \ours framework is able to achieve effective continual concept personalization, not only preventing the catastrophic forgetting but also preserving the zero-shot capability of pretrained text-to-image diffusion models.

Our contributions can be summarized as follows:
\begin{itemize}
    \item We present \ours, which advances neuron selection with an incremental finetuning strategy for continual personalization.
    \item We introduce a unique process for \concept selection, which identifies neurons related to target concept images and distinguishes them from the irrelevant ones during continual learning. 
    \item A incremental finetuning scheme is proposed with the selected \concept, which alleviates catastrophic forgetting while maintaining the zero-shot generation ability of the diffusion model.
    \item \ours is a fusion-free method for continual personalization. No additional model weights are needed to be stored, and no test-time optimization is needed either.
\end{itemize}

\section{Related Work}
\label{sec:related}

\subsection{Diffusion model personalization}

\paragraph{Single and multiple-concept personalization.}
Concept personalization~\citep{li2024blip, chen2023anydoor, zhang2024attention, motamed2023lego, kim2024selectively, gal2023encoder} seeks to adapt pretrained diffusion models for synthesizing personalized concepts with only a few example images. Some common concepts are subjects, background or style. Previous works \citep{gal2022image, ruiz2023dreambooth, voynov2023p+} achieved single concept personalization with several different strategies. For instance, ~\citet{gal2022image} proposes to enables LDMs to learn personalized concepts through optimization of new textual embeddings. In the meantime, ~\citet{ruiz2023dreambooth} assigns the user-specific concept to an unique identifier and finetunes the whole model. However, naively applying these methods to compose concepts in a single image usually results in overfitting and attributes binding. \citet{kumari2023multi} first proposes the joint-optimizing strategy and makes it realize to compose multi-concept in one image. Several following works~\citep{po2024orthogonal, gu2024mix, kumari2023multi, kwon2024concept, ding2024freecustom} tried to make improvement in this field but still suffer from some common issues such as over-fitting on the target images and zero-shot ability of text-to-image degradation as mentioned in~\cref{sec:intro}.

Low-Rank Adaption (LoRA)~\citep{hu2021lora} finetuning shows the capability of achieving down-stream tasks while preventing overfitting. Thus, to mitigate the aforementioned overfitting issue to this topic, a common solution is learning new concepts with LoRA~\citep{hu2021lora} finetuning. Previous works \citep{huang2025videomage, yang2024lora, po2024orthogonal, gu2024mix, kong2025omg, meral2024clora, wu2024mixture, shah2025ziplora} propose different strategies to merge the learned LoRA weights to generate multi-concept images. For example, \citet{gu2024mix} proposes gradient fusion method, which extracts all input and output features from each personalized LoRA. Gradient fusion method aims to ensure that the output features of the fused model closely match the output features of all individual personalized LoRAs. However, since this method requires storing all input and output features during the fusion process, it demands significant memory space and computational time. Instead of fusing LoRAs after learning each concept, \citet{po2024orthogonal} try to solve the problem from the initialization of LoRA weight. Under their assumption, they argue that as long as the down sampling part of LoRA weight across each concept are orthogonal to each other, then these LoRA weights keep independent and will not affect each other. However, as the number of concepts grows, the deviation of their assumption grows as well, which indicates the uncertainty of their method when fusing multiple personalized LoRA weights. Unfortunately, these aforementioned methods suffer from fusing multiple personalized LoRA, which also waste a lot of memory space and processing time.   

\paragraph{Continual concept personalization.}
Continual concepts personalization further aims to incrementally extend the learned concepts in a single latent diffusion model. \citet{smith2023continual} propose a self-regularization loss between LoRA weights in different stage to preserve previous learned concepts. Nevertheless, as the number of concepts grows, this proposed regularization method lost its function since the new concept LoRA weight will face the restriction from all of the previous learned LoRA weights. \citet{sun2024create} generate and store the data corresponding to the learned personalized concepts in a memory bank and wisely select images in the bank by their proposed rainbow-memory back strategy. Obviously, storing all of the relevant data of the learned personalized concepts is not realistic as the number of the concepts increases. \citet{dong2024continually} propose the task-shared project matrix to extract shared semantic information across different concepts and elastic weight aggregation during inference to tackle catastrophic forgetting. However, shared semantic information only exists between similar concepts, and elastic weight aggregation requires test-time optimization, which requires a lot of computational resources.  

\begin{figure*}[]
    \centering
    \includegraphics[width=\textwidth]{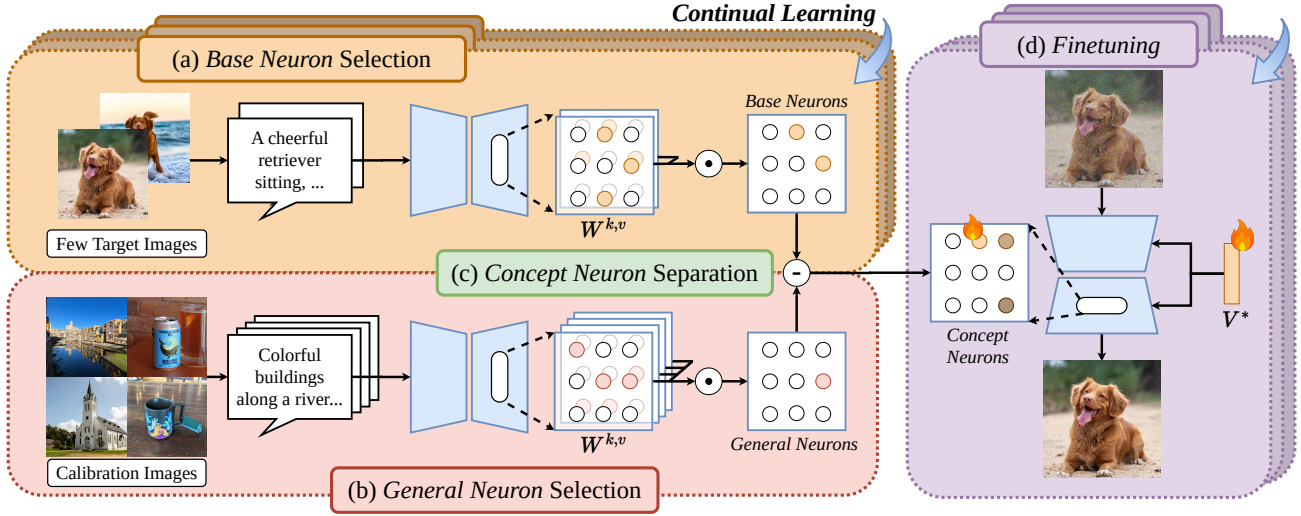}
    \caption{\textbf{Overview of \ours.} The proposed framework for neuron selection consists of (a) base neuron selection, (b) general neuron selection, and (c) concept neuron separation. With sparse, concept-specific neurons automatically selected fro each concept, the proposed incremental finetuning scheme in (d) update the text-to-image diffusion model for continual personalization.}
    \label{fig:main_figure}
\end{figure*}

\subsection{Neuron selection}
As the number of parameters in large language models (LLMs) rapidly increases, how to efficiently finetune these LLMs has become a significant issue. Research~\citep{wang2022finding, suau2020finding, dalvi2019one, durrani2020analyzing, antverg2021pitfalls} has provided evidence that certain neurons within feed-forward layers of transformer-based LLMs are intricately tied to task-specific outputs. Adjusting these neurons can significantly influence task performance, reducing the necessity for extensive finetuning. Building on this understanding, recent methods have been developed to detect modular structures within pretrained transformers, capitalizing on neuron sparsity~\citep{zhang2021moefication}. Moreover, studies reveal that these modules possess specialized functions, each serving unique roles within the model~\citep{zhang2023emergent}.

After many works on LLMs prove the effectively of neuron selection and its widely usage, \citet{liu2023cones} first applied this concept in the text-to-image diffusion model generation and customization. However, \citep{liu2023cones} has been proved that suffering from high training costs and low success rates along with the increased number of subjects in~\citep{liu2023cones2}, who needs an additional user-defined layout to improve it. Additionally, \citet{chavhan2024conceptprune} builds upon this foundation, explores by identifying neurons responsible for generating undesired concepts within diffusion models. Unlike in LLMs, isolating such neurons in LDMs presents unique challenges due to the complex aggregation across multiple denoising steps and the model’s sensitivity to intermediate outputs from prior time steps. Inspired by~\citep{chavhan2024conceptprune}, we propose~\ours to select sparse and highly related neurons in diffusion models to excel at the continual personalization.
\section{Preliminary of Diffusion Models}

\label{sec:preliminaries}
Text-to-image diffusion models generate images by progressively removing noise from an initial noisy input, guided by a conditioning vector derived from a text prompt. These models integrate the conditioning vector through cross-attention layers, which adjust the network’s latent features according to the text conditioning. This approach enables the model to produce outputs that accurately align with the prompt, allowing for precise control over the generated image's content. In cross-attention layers, let $\mathbf{c} \in \mathbb{R}^{s \times d}$ represent the text conditioning features with $s$ as the number of tokens and $d$ as the dimension of tokens. We also make $\mathbf{f} \in \mathbb{R}^{(h \times w) \times l}$ represent the latent image features. A single-head cross-attention operation \citep{vaswani2017attention} computes $Q = \mathbf{f} W^q$, $K = \mathbf{c} W^k$ and $V = \mathbf{c} W^v$, followed by a weighted sum:
\begin{equation}
\label{eq:attention}
\text{Attention}(Q, K, V) = \text{softmax}\left(\frac{Q K^T}{\sqrt{d'}}\right) V,
\end{equation}
where, $W^q$, $W^k$, and $W^v$ project the inputs to the query, key, and value spaces, with $d'$ as the dimension for key and query outputs. The attention block output is then used to update the latent features. 

These models minimize a loss function that refines the denoising process to approximate the target image, typically formulated as follows:

\begin{equation}
\label{eq:diffusion}
    \Eb{\bx,\bc,\bepsilon,t}{w_t \|\hat\bx_\theta(\alpha_t \bx + \sigma_t \bepsilon, \bc) - \bx \|^2_2},
\end{equation}

\noindent where $\bx$ is the ground-truth image, $\bc$ is the conditioning vector obtained from the text prompt, and $\alpha_t, \sigma_t, w_t$ are parameters controlling the noise schedule at time $t$.



\section{Method}
\label{sec:method}

\subsection{Problem formulation and framework overview}
\label{sec:formulation}


In the continual learning scheme, we define the $m$ representing the index of current concept to learn, and only $N_m$ images associated with concept $m$ is presented during training. Additionally, the model finetuned on $m$-th concept is expected to preserve the knowledge previously learned from concepts $1:m-1$. In the following section, we omit $m$ notation if we are discussing about only one concept.

We propose \textbf{C}oncept \textbf{N}euron \textbf{S}election, \ours, to achieve continual personalization while alleviating catastrophic forgetting. We introduce a neuron selection method that identifies a compact set of concept-specific neurons, which realizes the above continual learning scheme. As depicted in~\cref{fig:main_figure} and described in~\cref{sec:neurons_selection}, our proposed method learns and distinguishes between neurons for concept personalization and image generation. As detailed in~\cref{sec:reg_loss}, only the neurons associated with the input concept need to be updated and regularized in the continual learning scheme. As confirmed in our experiments, our method clearly perform against existing approaches for single and multi-concept image personalization.


\subsection{Learning of concept neurons}
\label{sec:neurons_selection}


\paragraph{Base neuron selection.}
Recent works on concept editing and model pruning~\citep{chavhan2024conceptprune, sun2023simple} note that neurons with large responses to the objective imply higher contributions to the learned model, and thus they are preferred to be updated/edited during the training stage. Following the observation in~\citep{kumari2023multi} that the cross-attention layer parameters have relatively higher correlations to the image personalization objective, we thus focus on neurons of cross-attention layers in diffusion models for neuron selection. 

To be more precise, we denote the weights of key and value mapping by $W^k \in \mathbb{R}^{d \times d_{k}}$ and $W^v \in \mathbb{R}^{d \times d_{v}}$, while the inputs of both mapping are text embedding denoted by $\mathbf{c}$. We obtain $\mathbf{c}$ text embedding from the concept image caption generated by a pretrained image captioning model~\citep{achiam2023gpt}. To assess the significance of each element in both weights $W^{k,v}$for a single image, we calculate the element-wise product of its magnitude with the $\ell_2$ norm of the text embedding feature $\mathbf{c}$ across dimensions following~\citep{sun2023simple}. 

Take $W^k$ as an example, we calculate the importance scores as follows:
\begin{equation}
\label{eq:wanda}
    \mathbf{S}(W^{k}, \mathbf{c}) =\left|W^{k}\right| \odot \big(\mathbf{1} \cdot \left\|\mathbf{c}\right\|_2 \
\big),
\end{equation}
\noindent where \( |\cdot| \) represents the absolute value, \( \|\mathbf{c}\|_2 \) denotes the \( \ell_2 \) norm applied to each column of \( \mathbf{c} \), producing a vector of dimension \( d \), and the symbol \( \odot \) indicates element-wise matrix multiplication. Note that in~\cref{eq:wanda}, $\mathbf{1} \cdot \left\|\mathbf{c}\right\|_2$ specifically uses broadcasting to apply \( \|\mathbf{c}\|_2 \) across the rows of \( W \), allowing for element-wise multiplication in each row. For a given row \( W_{i,:} \) with the associated importance scores \( \mathbf{S}(W, \mathbf{c})_{i, :} \in \mathbb{R}^{d \times d_{k}} \). We would select a neuron as a base one, if its score is above a pre-determined threshold (see details in the supplementary material).

Since there are $N$ images for concept, the above process would select neurons for each image, which can be viewed as a binary neuron mask $\bm{M}_{n}, n \in \{1,2,...,N\}$. We then aggregate all neuron masks obtained across images of the same concept and apply a logical \texttt{and} operation. We refer to such aggregated neurons as \base, as depicted in~\cref{fig:main_figure}(a). The resulting \base mask $\bm{M}^{base}$ can be expressed as:
\begin{equation}
\label{eq:important_mask}
    \bm{M}^{base} = \bigwedge_{n} \bm{M}_{n}, n \in \{1,2,...,N\}.
\end{equation}
\noindent Note that the above process is applied to both key and value mappings $W^{k,v}$ in all cross-attention layers.

\begin{figure}[t]
    \includegraphics[width=0.9\linewidth]{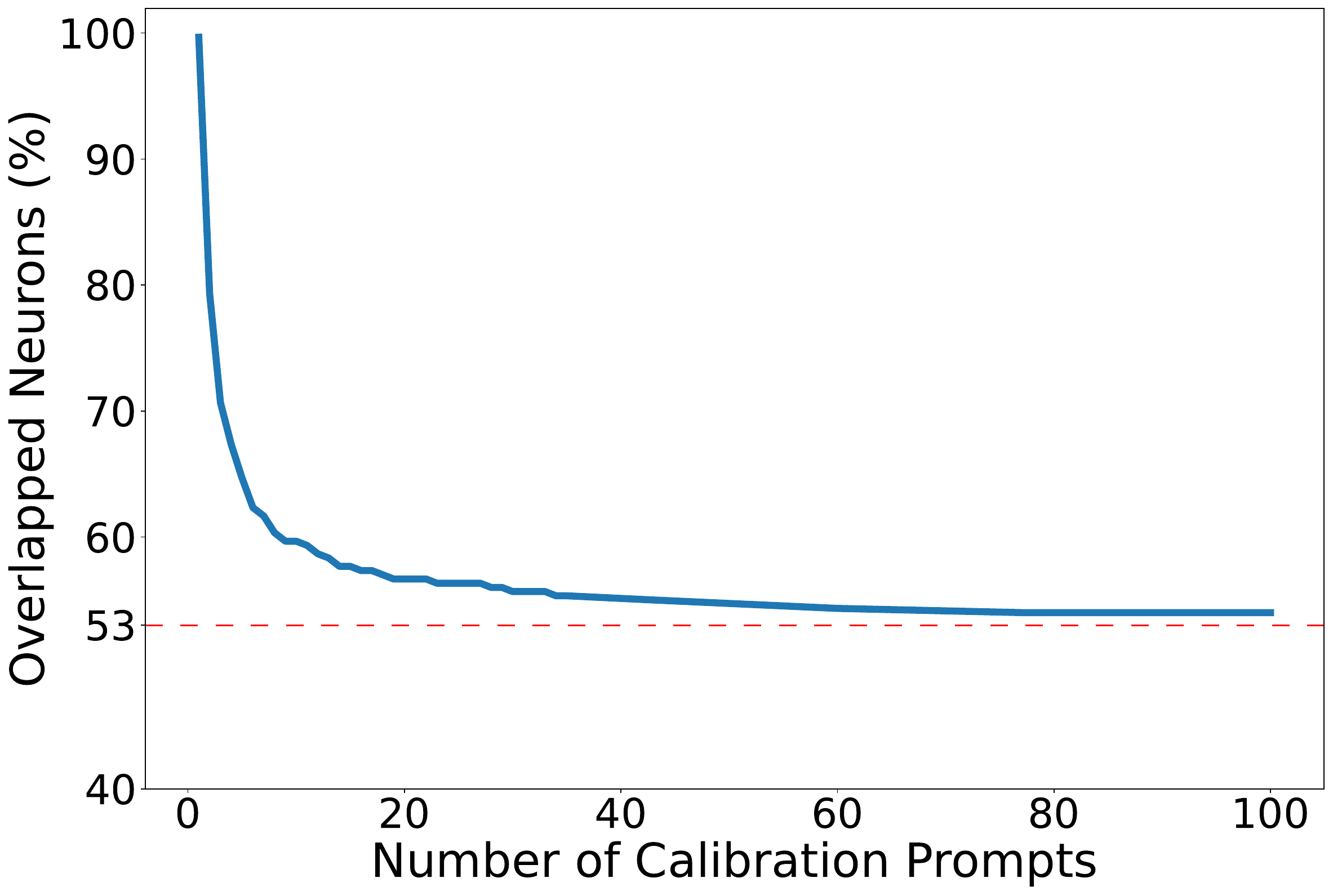}
    \caption{\textbf{Overlapped percentage of \base across images.} By increasing the number of text prompts, we observe a high percentage (around 53\%) of \base shared across the resulting images. This suggests that a large portion of \base share the goal of image generation, not concept personalization.}
    \label{fig:core_neurons}
\end{figure}

\textbf{}
\paragraph{General neuron selection.}
Despite the idea of selecting responsive neurons from concept images aligns with that of model pruning~\citep{chavhan2024conceptprune, sun2023simple}, we observe a large overlap between neurons selected from the image diffusion model when different text prompts are served as the inputs. As shown in~\cref{fig:core_neurons}, by increasing the number of input prompts as detailed in supplementary, we empirically observe that about $53\%$ of neurons are always selected. This suggests the aforementioned scheme not only selects neurons associated with the input prompt/condition but also identifies neurons contributing to the general image generation process. In \ours, we define the latter type of neurons as \general, and we aim to disregard such neurons when performing continual concept personalization. 

We now discuss how we detect general neurons which are related to image generation but not describing the concept of interest. As depicted in~\cref{fig:main_figure}(b), we collect a list of diverse calibration prompt set $\mathbf{P}_k$ for $k \in \{1, 2, \dots, K\}$ to find out \general in latent diffusion models' cross-attention layers. We use $K=20$ different prompts as calibration prompts to obtain the \general mask $\bm{M}^{general}$, which is obtained by:
\begin{equation}
\label{eq:core_mask}
    \bm{M}^{general} = \bigwedge_{k} \bm{M}_{k}, k \in \{1,2,...,K\}.
\end{equation}

\paragraph{Identification of concept neurons.}
With the collection of both \base and \general, it comes straightforward to exclude the \general from the \base for identifying the \concept of interest. This is simply achieved by performing a logical \texttt{not} operation on $\bm{M}^{base}$ and $\bm{M}^{general}$, resulting in the \concept mask $\bm{M}^{concept}$ for a specific concept. That is,
\begin{equation}
\label{eq:concept_mask}
    \bm{M}^{concept} = \bm{M}^{base} \land \neg\bm{M}^{general},
\end{equation}
\noindent where $\land$ and $\neg$ denote the logical \texttt{OR} and \texttt{NOT} operators. Again, we apply the aforementioned operation on both $W^{k,v}$ in all cross-attention layers. For each concept of interest, the derived \concept mask will be utilized for later continual learning.
It is worth noting that even though we subtract general neurons from base one, those general neurons are fixed (not pruned) during training. Therefore, the associated pre-trained ability would not be affected.

\begin{figure*}[t]
    \centering
    \includegraphics[width=0.95\textwidth]{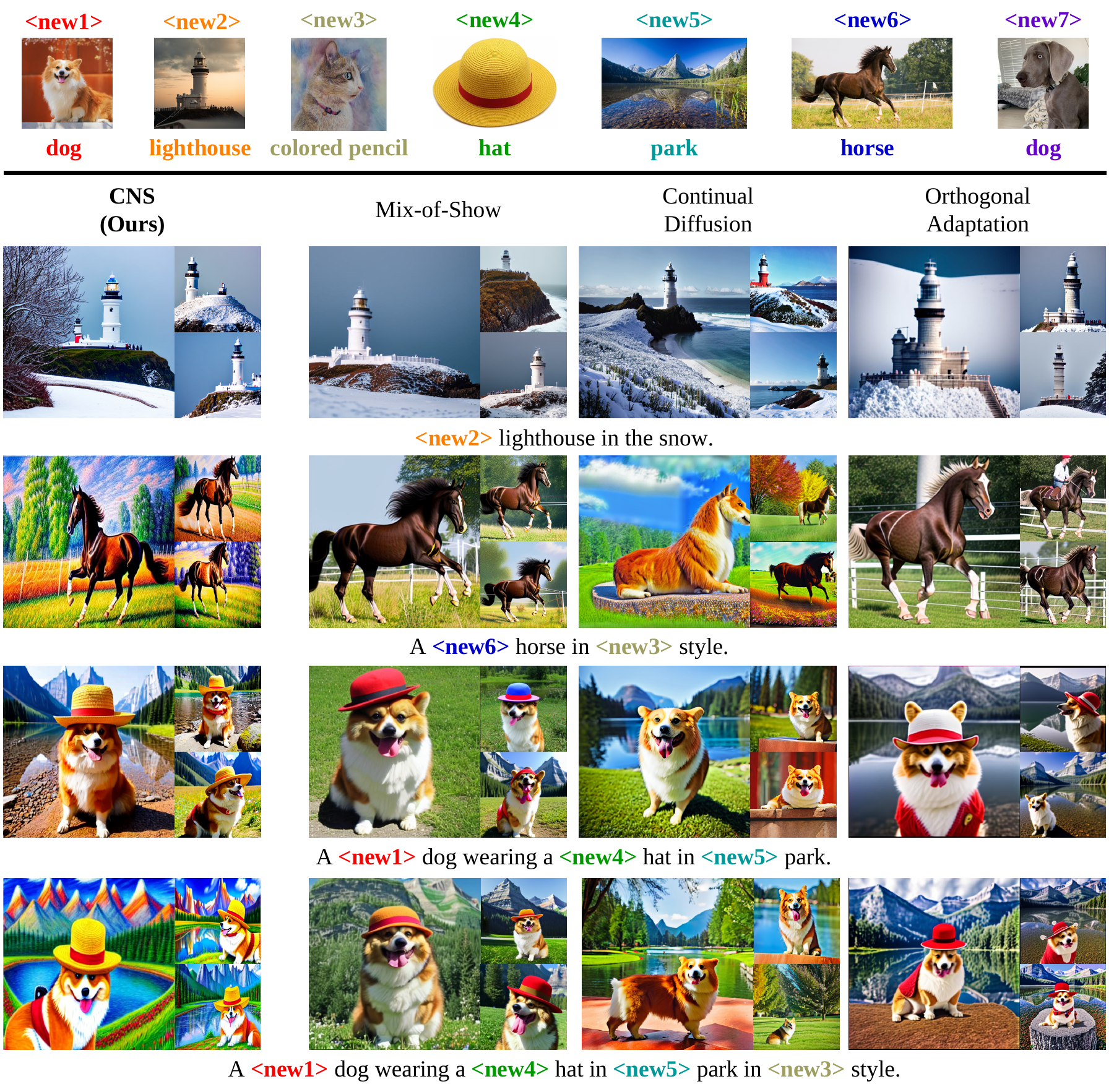}
    \caption{\textbf{Qualitative visualization.} Note that only Continual Diffusion~\citep{dong2024continually} and \ours are capable of performing continual personalization, while Mix-of-Show~\citep{gu2024mix} and Orthogonal Adaptation~\citep{po2024orthogonal} require to keep LoRAs for each concept for personalization. It can be seen that our personalized outputs match concepts learned across different time, alleviating appearance leakage and catastrophic forgetting problems.}
    \label{fig:comparison_figure}
\end{figure*}


\subsection{Continual personalization with concept neurons}
\label{sec:reg_loss}

With \concept for each concept properly identified, the final challenge is to perform continual concept personalization while mitigating possible catastrophic forgetting problems. Since there is no guarantee that \concept of different concepts are distinct, we propose to perform continual learning with \textit{neuron regularization}:
\begin{multline}
\label{eq:reg_loss}
    L_{reg} = \lambda_1 \|W_m \odot \bm{M}^{reg} - W_{m-1} \odot \bm{M}^{reg}\|_2 + \\
    \lambda_2 \|W_m \odot \bm{M}^{reg} - W_{0} \odot \bm{M}^{reg}\|_2, 
\end{multline}
\noindent where $\bm{M}^{reg} = \bm{M}^{concept}_{m} \land (\bigvee_{m'=1}^{m-1} {M}^{concept}_{m'})$ denotes the regularization neuron masks, which indicates the intersection of the neurons to be updated and the neurons have been tuned for the previously learned concepts. $W_m$ indicates the model weights of the neurons to be updated when learning the $m^{\text{th}}$ concept, $W_{m-1}$ indicates the model weights after customized on the $m-1^{\text{th}}$ concept and $W_0$ indicates the pretrained weights of the original diffusion model. In~\cref{eq:reg_loss}, the first term regularizes the model weights previously learned to prevent catastrophic forgetting (i.e., prior personalized concepts). On the other hand, the second term in~\cref{eq:reg_loss} regularizes the model weights with the pretrained ones to prevent zero-shot capability degradation.

Combining with the loss function presented in~\citep{ruiz2023dreambooth} to prevent from subject overfitting, our proposed framework is capable of retaining both previously learned concepts from $W_{m-1}$ and the inherent zero-shot text-to-image capability from the pretrained diffusion model $W_0$ during finetuning on new concept. In summary, the overall objective function of our proposed continual personalization loss via concept neurons selection is formulated as below:
\begin{multline}
\label{eq:condense_loss}
    \E_{\bx,\bc,\bepsilon,\bepsilon^\prime,t}[w_t \|\hat\bx_\theta(\alpha_t \bx + \sigma_t \bepsilon, \bc) - \bx \|^2_2 + \\
    \lambda w_{t^\prime} \|\hat\bx_\theta(\alpha_{t^\prime} \bx_\text{pr} + \sigma_{t^\prime} \bepsilon^\prime, \bc_\text{pr}) - \bx_\text{pr} \|^2_2 + L_{reg}].
\end{multline}

While training, we update the \concept and special prompt token following~\citep{kumari2023multi} by minimizing~\cref{eq:condense_loss}, without the need to store any additional model weights during continual learning. For inference, \ours can be directly applied to produce images with concepts observed anytime during training, users do not spend any additional computational effort such as test-time optimization.
\section{Experiment}
\label{sec:exp}
\begin{table*}[t]
\caption{\textbf{Quantitative comparisons of single and multi-concept personalization.} In addition to the alignment-based metrics of CLIP-I and CLIP-T, we provide the computation estimates for different personalization methods. Note that memory requirements for GPU/CPU and computation time for multi-concept personalization indicate the \textit{additional} costs for fusing concept weights previously learned.}
\centering
\newcolumntype{C}{>{\centering\arraybackslash}X}
\newcolumntype{L}{>{\raggedright\arraybackslash}X}
\newcolumntype{R}{>{\raggedleft\arraybackslash}X}
\begin{tabularx}{\textwidth}{lCCCCCC}
\toprule
\multirow{3}{*}{Methods} & \multicolumn{2}{c}{Single Concept} &  \multicolumn{2}{c}{Multiple Concepts} & \multicolumn{2}{c}{Computational Resources} \\
\cmidrule(lr){2-3}
\cmidrule(lr){4-5}
\cmidrule(lr){6-7}
       & CLIP-I $\uparrow$ & CLIP-T$\uparrow$ & CLIP-I$\uparrow$ & CLIP-T$\uparrow$ & Memory(MB)$\downarrow$ & Time(s)$\downarrow$ \\
\midrule
Textual Inversion~\citep{gal2022image}   & 72.76 & 72.69 & 65.30 & 65.00 & \textbf{0} / \textbf{0} & \textbf{0} \\
Custom Diffusion~\citep{kumari2023multi}  & 67.88 & 74.92 & 65.87 & 68.70 & 3547 / \textbf{0} & 10 \\
Mix-of-Show~\citep{gu2024mix} & \textbf{75.86} & 75.75 & 65.26 & 70.62 & 62852 / \textbf{0} & 727 \\
Orthogonal Adaption~\citep{po2024orthogonal}  & 74.67 & 74.87 & 66.37 & 69.20 & 5663 / 3167 & 42 \\
Continual Diffusion~\citep{smith2023continual}  & 71.82 & 66.12 & 66.15 & 60.30 & 2461 / 4747 & 10 \\
\ours  & 74.88 & \textbf{76.95} & \textbf{67.21} & \textbf{79.22} & \textbf{0} / \textbf{0} & \textbf{0} \\

\bottomrule
\end{tabularx}
\label{tab:clip_score}
\end{table*}
\begin{table}[t]
\caption{\textbf{Ablation study of our approach.} We compare \ours with two baselines, removing the continual regularization loss in~\cref{sec:reg_loss} and randomly picking fine-tuned neurons (i.e., no \concept selection of~\cref{sec:neurons_selection}).} 
\centering
\newcolumntype{C}{>{\centering\arraybackslash}X}
\newcolumntype{L}{>{\raggedright\arraybackslash}X}
\newcolumntype{R}{>{\raggedleft\arraybackslash}X}
\begin{tabularx}{\columnwidth}{CCCCCCC}
\toprule
\multicolumn{3}{c}{Components} & \multicolumn{2}{c}{Single Concept}  \\
\cmidrule(lr){1-3}
\cmidrule(lr){4-5}
random & concept & \multirow{2}{*}{$L_{reg}$} & \multirow{2}{*}{CLIP-I$\uparrow$} & \multirow{2}{*}{CLIP-T$\uparrow$} \\
 neurons & neurons & & & \\

\midrule
\checkmark &  & \checkmark & 73.15 & 73.30 \\
 & \checkmark &  & 73.06 & 74.65 \\
 & \checkmark & \checkmark & \textbf{74.88} & \textbf{76.95} \\
\midrule
\multicolumn{3}{c}{} & \multicolumn{2}{c}{Multiple Concepts}  \\
\cmidrule(lr){1-3}
\cmidrule(lr){4-5}
\checkmark &  & \checkmark & 65.05 & 72.92 \\
 & \checkmark &  & 66.89 & 77.02 \\
 & \checkmark & \checkmark & \textbf{67.21} & \textbf{79.22} \\
\bottomrule
\end{tabularx}
\label{tab:reg_and_random}
\end{table}

\begin{figure}[t]
    \includegraphics[width=1\linewidth]{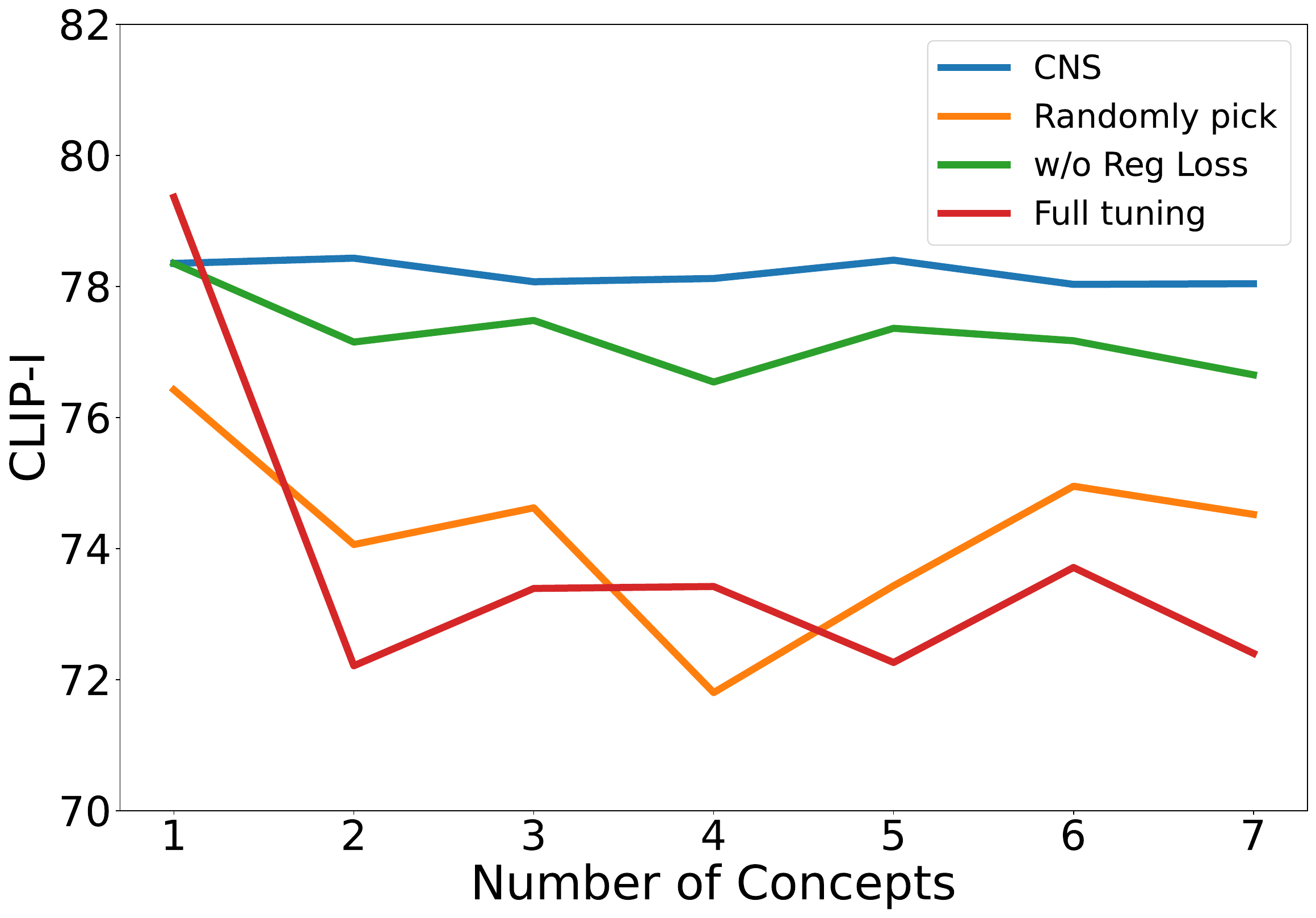}
    \caption{\textbf{Performance degradation on the first concept over time.} Compared to the ablated version of our method, the full version of \ours is sufficiently robust during continual learning, resulting in negligible degradation on the first concept. This confirms our ability in alleviating catastrophe forgetting problems.}
    \label{fig:forgetting}
\end{figure}

\subsection{Experimental setup}
\begingroup
\renewcommand\thefootnote{}
\footnotetext{The authors from NTU downloaded, evaluated, and completed the experiments on the datasets.}
\addtocounter{footnote}{-1}
\endgroup
\label{sec:experimental_setup}
\paragraph{Dataset.} We conduct experiments on a dataset containing $20$ concepts, composed with $8$ real-world animals, $6$ real-world objects, $3$ styles and $3$ real-world scenes. More details are provided in the supplementary.
\paragraph{Implementation details.}
We leverage Stable Diffusion (SD-1.5\footnote{https://huggingface.co/runwayml/stable-diffusion-v1-5}) as our pretrained text-to-image generation model to conduct comparison experiments. We fine-tune both the text embeddings and \concept using the Adam optimizer, with a learning rate of $5e-4$ for text embeddings and $3e-5$ for \concept. The training steps are $500$ for a single concept, which takes about four minutes on a single 3090 GPU. For all the other methods mentioned in this paper, we use the same backbone and pretrained weight as ours. For more implementation details, we provide them in the supplementary.

\paragraph{Evaluation metrics.}
Following~\citep{kumari2023multi}, we use CLIP~\citep{radford2021learning} to evaluate image- and text-alignment. For single-concept image alignment, both the generated and concept images are fed into the CLIP image encoder to obtain embeddings, and their cosine similarity is calculated. For multi-concept personalization, image alignment is measured as the average visual similarity between the generated image and each concept image. To assess text similarity, the CLIP image encoder processes the generated images and the CLIP text encoder processes the text prompt. The cosine similarity between these embeddings serves as the text alignment score.

\subsection{Qualitative comparisons}
We demonstrate the result of \ours and other competitive methods in~\cref{fig:comparison_figure}.  For Mix-of-show~\citep{gu2024mix} and Orthogonal Adaptation~\citep{po2024orthogonal}, we learned LoRA for each concept and merged them by their proposed fusion methods. For the others, we incrementally learn from the first concept to the seventh concept. As shown in~\cref{fig:comparison_figure}, \ours outperforms all the other methods in both single-concept and multi-concept personalization. For single-concept setting, \ours is able to match the prompt perfectly and consistently. As more concepts are included in prompts, the other three methods suffer from concept vanish or attribute binding (\eg Continual Diffusion generates a $<$new1$>$ dog-like horse even though $<$new1$>$ dog is not mentioned in the prompt.). Yet, images generated by \ours still preserve the image-alignment for all concepts and text-alignment for input prompt.

\subsection{Quantitative comparisons}
\label{sec:quantitative}
\paragraph{Quantitative comparisons of image quality.}
We evaluate $3$ continual personalization sets, each set contains $7$ concepts including animals, styles, objects and scenes. Each set contains $35$ single-concept prompts, $7$ multi-concept prompts. We generate $50$ images per prompt and pick the best as representative image. 

As shown in the first column of~\cref{tab:clip_score}, \ours is slightly lower than Mix-of-Show~\citep{gu2024mix} on single-concept image-alignment yet attains the highest score on single-concept text-alignment, which indicates that \ours does not suffer from overfitting the concept image, and it effectively captures semantic details from text descriptions. \ours outperforms all the other method in multi-concept image-alignment and text-alignment. It is worth mentioning that all the other methods suffer from performance degradation on multi-concept text-alignment compared to single-concept. In the contrast, \Cref{tab:clip_score} shows that \ours preserves text-to-image ability to multi-concept generation and achieve the highest image quality.

\paragraph{Quantitative comparisons of computational efficiency.}
\Cref{tab:clip_score} compares memory and time consumption across various fusion methods when fusing seven concepts. Notably, while Mix-of-Show~\citep{gu2024mix} achieves near-top scores in both CLIP-I and CLIP-T, it also demands the highest memory and time resources, highlighting a trade-off between fusion efficiency and performance in prior methods. In contrast, \ours demonstrates balanced fusion efficiency and high performance, eliminating the need for such compromises. Additionally, to avoid concept fusion issues, methods like~\citep{gu2024mix, po2024orthogonal} combine LoRA weights for each concept within a single image, though this requires multiple fusion operations for different concept combinations.

\subsection{Ablation Study}
\label{sec:ablation}
\paragraph{Quantitative ablation study.}
In \cref{tab:reg_and_random}, we examine the effects of regularization loss and targeted fine-tuning of concept neurons. To assess the impact of fine-tuning \concept, we randomly selected the same number of neurons as \concept and fine-tuned them. As shown in the first row of \cref{tab:reg_and_random}, both CLIP-I and CLIP-T scores decrease in single- and multi-concept scenarios, indicating that our neuron selection method successfully identifies the most representative neurons, enhancing personalization performance. Additionally, we ablate the regularization loss $L_{reg}$, which prevents catastrophic forgetting by preserving overlapping concept neurons with previous concept and pretrained weights. Without $L_{reg}$, some overlapping neurons are overwritten by new concepts, leading to catastrophic forgetting and degraded performance.

\paragraph{Catastrophic forgetting ablation study.}
To prove that our \ours framework is robust enough for preventing from catastrophic forgetting, we conduct an experiment as depicted in~\cref{fig:forgetting}. We illustrate the degradation of CLIP-I score of the first learned concept after learning multiple concepts incrementally. In this experiment, we report the degradation curves of the four baselines: the green curve represents ours method without continual regularization loss as detailed in~\cref{sec:reg_loss}; the yellow curve represents that we fine-tune same amount of randomly picked neurons as our selected \concept as detailed in~\cref{sec:neurons_selection}. The red curve represents fine-tune all the parameters of key and value mapping from the text latent in the cross-attention layers. The results show that with \concept selection and the continual regularization loss, \ours can realize continual personalization while preventing from catastrophe forgetting effectively comparing to all the baselines.

\paragraph{Human evaluation.}
\begin{table}[t]
\caption{\textbf{Human study.} Scores for image and text alignments denote the percentages of users who considered the image generated by the corresponding method to be the most desirable.}
\centering
\newcolumntype{C}{>{\centering\arraybackslash}X}
\newcolumntype{L}{>{\raggedright\arraybackslash}X}
\newcolumntype{R}{>{\raggedleft\arraybackslash}X}
\begin{tabularx}{\columnwidth}{lCC}
\toprule
\multirow{2}{*}{\textbf{Methods}} & Image Alignment($\%$) & Text Alignment($\%$) \\
\midrule
Textual Inversion~\citep{gal2022image} & 3.1 & 4.7 \\
Custom Diffusion~\citep{kumari2023multi}  & 1.2 & 0.8  \\
Mix-of-Show~\citep{gu2024mix} & 20.2 & 22.3  \\
Orthogonal Adaption~\citep{po2024orthogonal}  & 14.3 & 16.2  \\
Continual Diffusion~\citep{smith2023continual}  & 1.4 & 0.9  \\
\ours  & \textbf{59.8} & \textbf{55.1} \\
\bottomrule
\end{tabularx}
\label{tab:human}
\end{table}
We hire 10 users to conduct human study to further evaluate \ours. In the experiment, we provide 5 single-concept and 15 multi-concept prompts, and generate 5 samples for each prompt. For each question, users are required to select the image aligned mostly with the input prompt and the image most similar to the target concepts' images across all samples generated by each method. The results are shown in~\cref{tab:human}.

\vspace{1mm}
\paragraph{Further analysis.}
We present additional results in the supplementary. For example, we conduct experiments to confirm our scoring mechanism, and those verifying the selected neurons during continual personalization. And, experiments with region control show that \ours can be integrated with different text-to-image generation models. 

\section{Conclusion and Future Work}
\label{sec:conclusion}

We present Concept Neuron Selection, \ours, a simple yet effective approach for continual learning-based personalization. By identifying concept-specific neurons in LDMs, \ours incrementally fine-tunes these neurons while preserving prior knowledge and zero-shot text-to-image generation. \ours achieves state-of-the-art performance with minimal parameter updates, outperforming existing methods in single- and multi-concept scenarios. Additionally, \ours operates fusion-free, reducing memory and processing demands for continual personalization.

Compared to previous continual personalization approaches which require to store LoRAs for each concept, we only need to store the concept neuron masks $\mathbf{M}_{1:m}$ to track previously tuned neurons, with significantly less memory requirement. As future research directions, our proposed concept neuron selection scheme can be possibly extended to tackle knowledge editing and unlearning tasks, not limited to particular data modality. 

\section{Acknowledgment}
This work is supported in part by the National Science and Technology Council via grant NSTC113-2634-F-002-005, and in part by Qualcomm Technologies, Inc. through a Taiwan University Research Collaboration Project.
{
    \small
    \bibliographystyle{ieeenat_fullname}
    \bibliography{main}
}
\clearpage
\setcounter{page}{1}


\twocolumn[{
\renewcommand\twocolumn[1][]{#1}
\maketitlesupplementary
\begin{center}
    \centering
    \includegraphics[width=\textwidth]{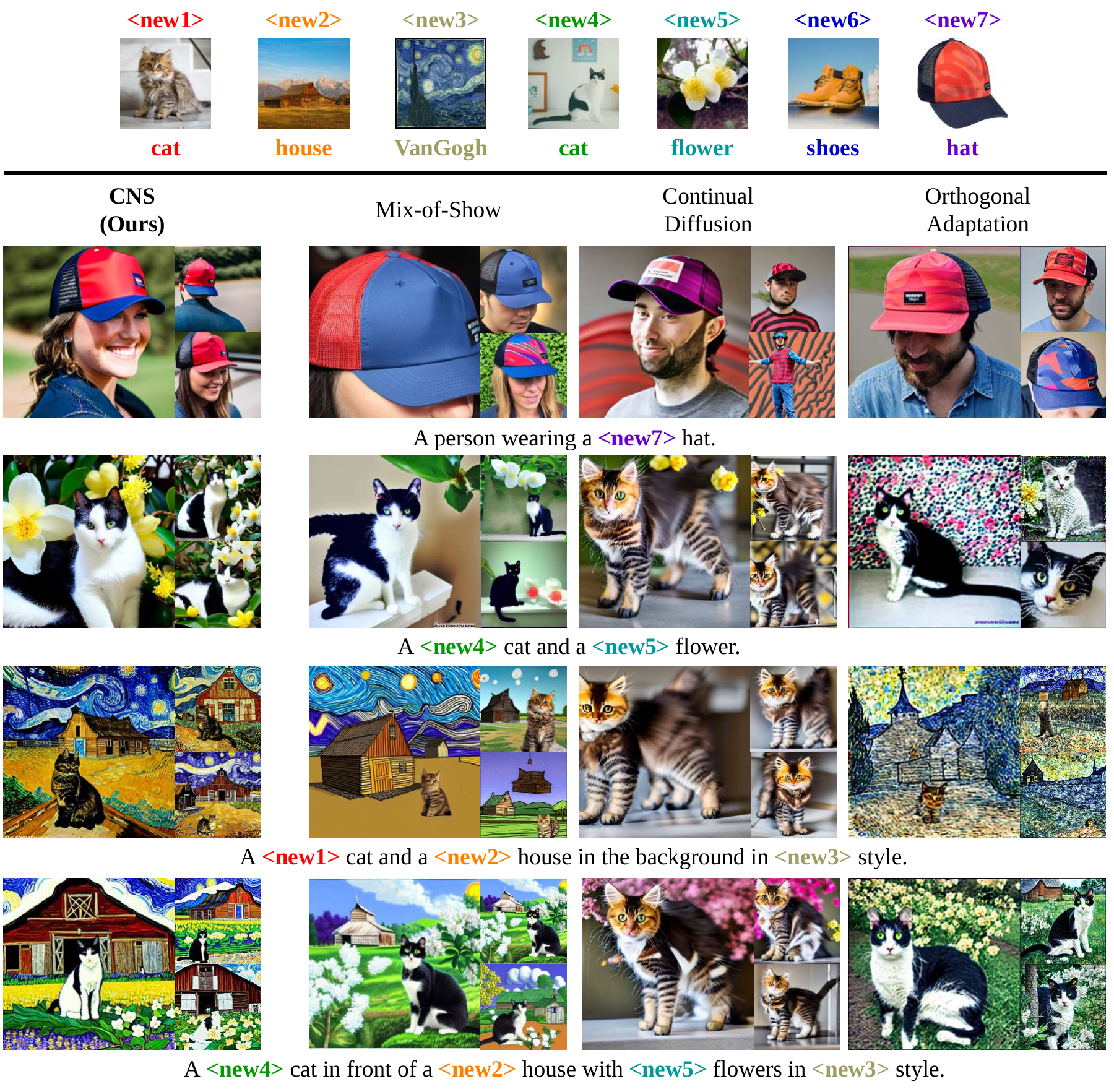}
    \captionof{figure}{\textbf{More qualitative visualization.} Note that only Continual Diffusion~\citep{dong2024continually} and \ours are capable of performing continual personalization, while Mix-of-Show~\citep{gu2024mix} and Orthogonal Adaptation~\citep{po2024orthogonal} require to keep LoRAs for each concept for personalization. It can be seen that our personalized outputs match concepts learned across different time, alleviating appearance leakage and catastrophic forgetting problems.}
    \vspace{9mm}
\label{fig:teaser}
\end{center}
}]
\begin{figure}[t]
    \includegraphics[width=\linewidth]{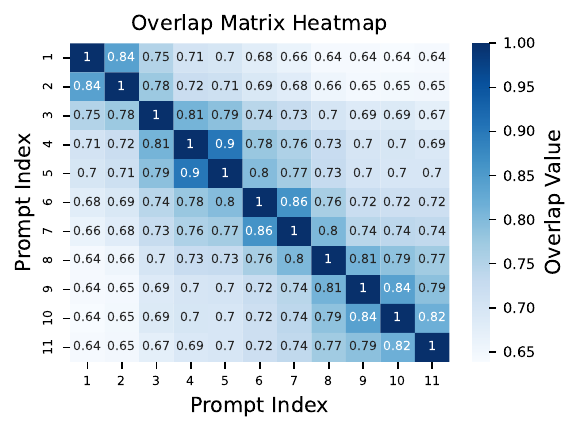}
    \caption{Confusion matrix of the similarity scores between query pairs. This figure was provided in Supplementary Sec. 2.}
    \label{fig:overlap_matrix_heatmap}
\end{figure}

\begin{figure*}[t]
    \centering
    \includegraphics[width=0.9\textwidth]{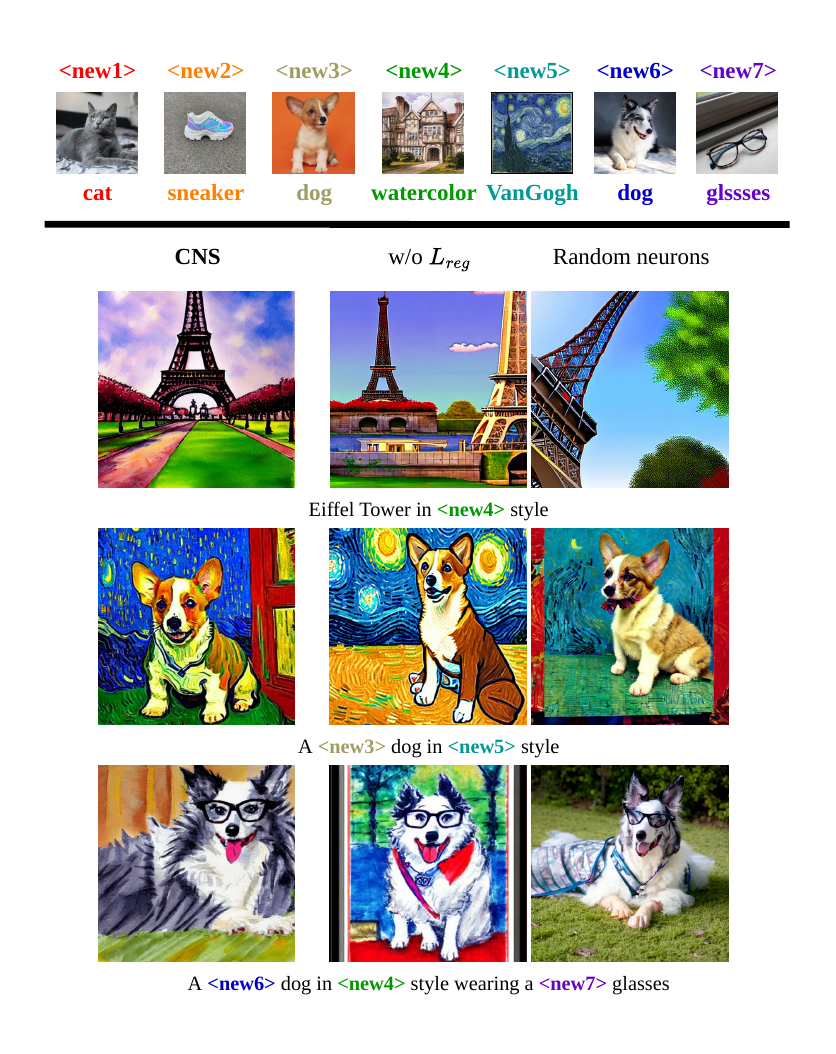}
    \caption{\textbf{Qualitative result of ablation study.} It was observed that without the assistance of $L_{reg}$, the objects in the images lost detail due to catastrophic forgetting. Additionally, when the same number of neurons as the concept neurons are randomly selected, the model struggles to learn the target concepts effectively, leading to noticeable performance degradation.}
    \label{fig:ablation}
\end{figure*}
\begin{figure*}[t]
    \centering
    \includegraphics[width=0.9\textwidth]{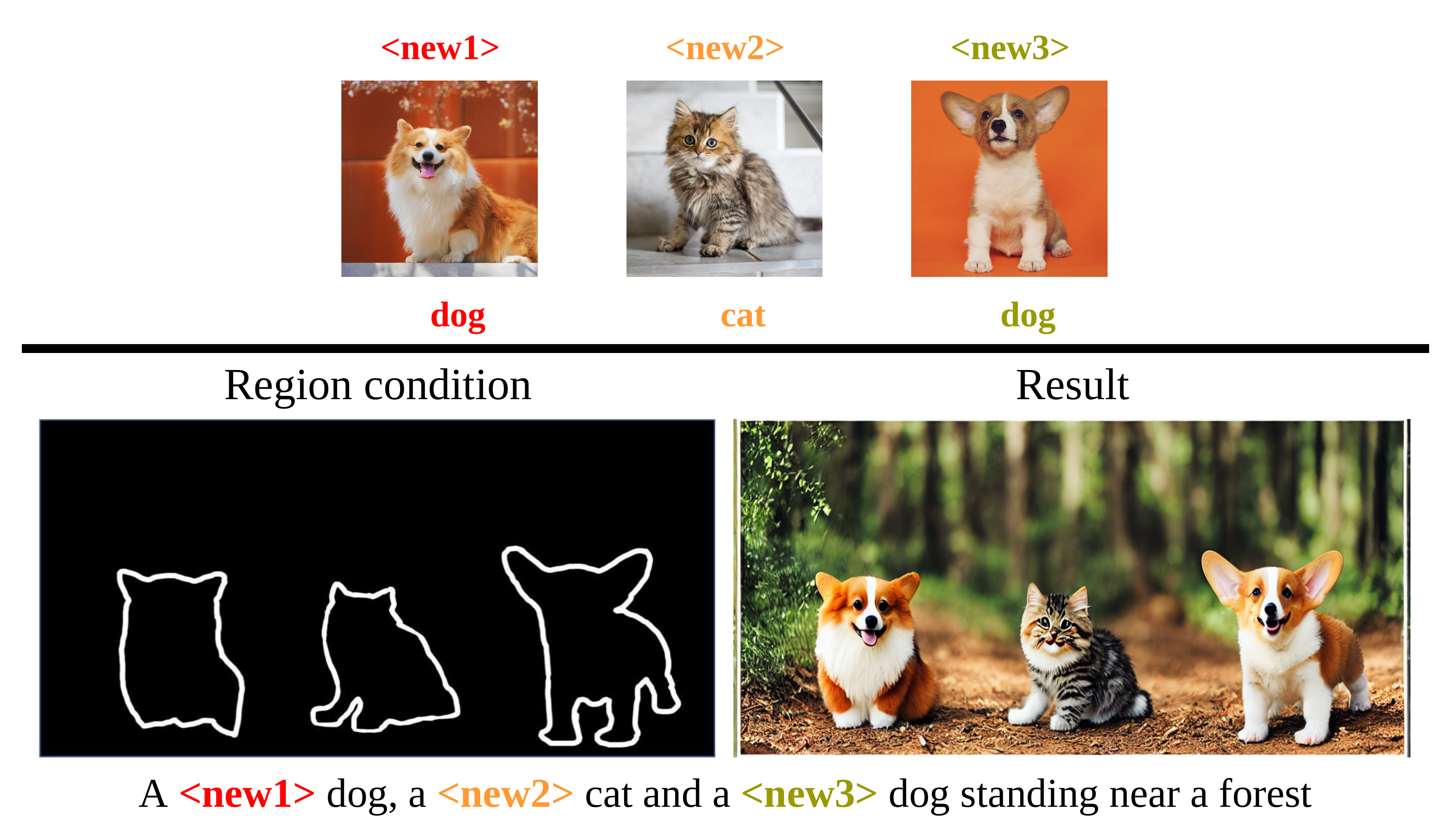}
    \caption{\textbf{Region control experiment.} In this figure, we shows the result of applying region control with \ours. Aside from fusing different concepts into an image, \ours can easily apply any test-time region control to specify where each concept should be placed.}
    \label{fig:region_control}
\end{figure*}

\section{Implementation Details}
\label{sec:implementation_details}
\subsection{Datasets}
To evaluate the continual personalization, we collect $3$ sets of concept images from~\citep{ruiz2023dreambooth, shah2025ziplora} and open source images on the net\footnote{https://yuhhuey1.pixnet.net/blog/post/221501409}. All set includes $7$ concepts and $42$ prompts for the evaluation. There are $35$ prompts with single concepts, $4$ prompt with double concepts and $3$ prompts with triple concepts, totally $42$ prompts. Concept types in the first set include three animals, two objects and two styles. Concept types in the second set include three animals, two background scenes, one object and one style. Concept types in the third set include two animals, one background scene, three objects and one style. We evaluate ours framework along with all the other baselines with the same dataset setting.

\subsection{Baseline methods}
\paragraph{Textual Inversion}
We clone the code base of textual inversion from official GitHub repository\footnote{https://github.com/rinongal/textual\_inversion}. The learning rate of text embedding is $5e-4$ while training steps is set to $1500$. For multi-concept personalization, we naively feed all the trained special tokens to the text-to-image diffusion model.

\paragraph{Custom Diffusion}
We clone the code base of custom diffusion from official GitHub repository\footnote{https://github.com/adobe-research/custom-diffusion}. The learning rate of both text embedding and diffusion model are $4e-5$ while training steps is set to $2500$. In the experiment of multi-concept personalization presented in Tab.~1, we fuse all the learned model weights with constrained optimization to merge concepts as described in~\citep{kumari2023multi}.

\paragraph{Mix-of-Show}
We clone the code base of custom diffusion from official GitHub repository\footnote{https://github.com/TencentARC/Mix-of-Show}. The learning rate of text embedding is $1e-3$ and LoRA is $1e-4$ while batch size is set to $2$. Training steps across each concept are different and determined by the number of images of each concept, as the setting in official code. In the experiment of multi-concept personalization presented in Tab.~1, we fuse all the learned LoRA weights with LoRA fusion approach as described in~\citep{gu2024mix}.

\paragraph{Orthogonal Adaption}
We implement Orthogonal Adaption by ourselves because official code is not available during our research process. We use the recommended randomized orthogonal basis, which is consistent with the paper. The learning rate of text embedding is $1e-3$ and LoRA is $1e-5$ while training steps is set to 500. 

\paragraph{Continual Diffusion}
We also implement Continual Diffusion by ourself, as the official code is unavailable during our research.
We follow the self-regularization loss presented in~\cite{smith2023continual} to fulfill continual personalization. The learning rate for both text embedding and LoRA are $5e-4$ while the training steps is set to $700$.

\subsection{Threshold of the neuron selection}

We define a threshold of $30\%$ for the neuron selection process described in Sec.~4.2. Specifically, we select the top $30\%$ of representative neurons in each row based on the importance scores computed using Eq.~(3).

\section{Additional Experimental Results}
\label{sec:additional_exp_results}

\subsection{Use of similarity scores}
To verify this, we now conduct experiments with 11 text queries, which is constructed as follows. We start with a text query with 5 adjective-noun pairs; then, we replace an adjective or a noun to produce another query, until all words are replaced (as the 11th query). We list three examples:
\begin{enumerate}
    \item Rainy beach, orange sky, palm trees, green sand, calm ocean
    \item Rainy beach, ..., \textcolor{cyan}{wild} ocean
    \item Rainy beach, ..., \textcolor{cyan}{wild} \textcolor{blue}{horse}
\end{enumerate}
\noindent Note that larger similarity scores indicate a higher overlapping percentage of selected neurons. With the aforementioned 11 queries, \cref{fig:overlap_matrix_heatmap} shows the associated confusion matrix (i.e.,  query pairs with smaller differences in words show larger overlapping/similarity scores). With the above experiments, our neuron selection scheme for identifying the described concepts can be verified.

\subsection{Percentage of the updated neurons}
In this section, we provide a detailed calculation of the updated neuron percentage within our framework. In \ours, we target only the linear layers in the cross-attention components $W^{k,v}$, as described in Sec.~3. From these, we identify the top $30\%$ of the \base using the importance scores calculated by Eq.~(3). Additionally, we perform the \concept selection process to isolate the \concept from the \base, resulting in about
$5\%$ of the neurons remaining in the cross-attention layers. Ultimately, this process updates only around $0.13\%$ of the total parameters for a single concept personalization.

\subsection{Percentage of the overlapping between concept neurons}
In this section, we present an overall estimation of the overlapping percentage among \concept for different concepts. Specifically, we compute the pairwise overlapping mIoU between concepts, finding an average overlapping mIoU of $0.24$ across \concept. Across all $7$ concepts in one set, the overlapping mIoU is $0.02$.

\subsection{Empirical experiment of general neurons}
In this section, we provide the details of how we calculate \general. We prompt the GPT~\citep{achiam2023gpt} to generate $100$ diverse image descriptions and collect them to identify the \general as detailed in Sec.~4.2.

\subsection{Visualization of ablation study}
\Cref{fig:ablation} shows the qualitative result of our ablation experiments mentioned in Sec.~5.4. Noted that without the help of $L_{reg}$, the objects in the images lost details as the result of catastrophic forgetting. In the meantime, if we randomly pick the same amount of neurons as concept neurons, model fails to learn target concepts properly and degradation of performance can be easily observed.

\subsection{Region control}
\label{sec:attention_control}
We have also observed the notorious attribute binding issue in \ours. However, \ours can easily leverage any test-time region control method to solve attribute binding. In~\cref{fig:region_control}, we illustrate the result of combining \ours with the region control method proposed in~\citep{gu2024mix} to mitigate the issue of attribute binding.


\end{document}